%% file: main.tex
\documentclass[10pt,twocolumn,letterpaper]{article}

\usepackage[algorithms]{wacv}      

\input{preamble}

\definecolor{wacvblue}{rgb}{0.21,0.49,0.74}
\usepackage[pagebackref,breaklinks,colorlinks,allcolors=wacvblue]{hyperref}
\def\wacvPaperID{253} 
\def\confName{WACV}
\def\confYear{2026}

\title{Training-free Conditional Image Embedding Framework Leveraging\\Large Vision Language Models}

\author{Masayuki Kawarada \qquad Kosuke Yamada \qquad Antonio Tejero-de-Pablos \qquad Naoto Inoue\\
CyberAgent, Japan\\
{\tt\small \{kawarada\_masayuki, kosuke\_yamada, antonio\_tejero\}@cyberagent.co.jp}
}

\begin{document}
\maketitle

\input{sec/0_abstract}    
\input{sec/1_intro}
\input{sec/2_related_work}
\input{sec/3_proposed_method}
\input{sec/4_experimental_settings}

\input{sec/5_results}
\input{sec/6_conclusion}
\input{sec/7_acknowledgments}

{
    \small
    \bibliographystyle{ieeenat_fullname}
    \bibliography{main}
}

\input{sec/X_supplementary}

\end{document}

%% file: preamble.tex
\newcommand{\red}[1]{{\color{black}#1}}
\newcommand{\blue}[1]{{\color{black}#1}}

\usepackage{multicol}
\usepackage{multirow}
\usepackage{chngcntr}
\usepackage{graphicx}
\usepackage{amsmath}
\usepackage{amssymb}
\usepackage{booktabs}
\usepackage{caption}

%% file: sec/0_abstract.tex
\begin{abstract}
Conditional image embeddings are feature representations that focus on specific aspects of an image indicated by a given textual condition (\eg, color, genre), which has been a challenging problem.
Although recent vision foundation models, such as CLIP, offer rich representations of images, they are not designed to focus on a specified condition.
In this paper, we propose \textbf{DIOR}, a method that leverages a large vision-language model (LVLM) to generate conditional image embeddings.
DIOR is a training-free approach that prompts the LVLM to describe an image with a single word related to a given condition. 
The hidden state vector of the LVLM's last token is then extracted as the conditional image embedding.
DIOR provides a versatile solution that can be applied to any image and condition without additional training or task-specific priors.
Comprehensive experimental results on conditional image similarity tasks demonstrate that DIOR outperforms existing training-free baselines, including CLIP.
Furthermore, DIOR achieves superior performance compared to methods that require additional training across multiple settings.\footnote{Our implementation is available at \url{https://github.com/CyberAgentAILab/DIOR_conditional_image_embeddings}}

\end{abstract}

%% file: sec/1_intro.tex
\section{Introduction}
\label{sec:intro}

Image embedding extraction is a fundamental technique for transforming high-dimensional image data into a low-dimensional, more manageable numerical representation. The similarity between embeddings from two images, often referred to as image similarity, has been foundational for many applications in computer vision, such as image retrieval~\cite{radenovic2018revisiting,revaud2019learning,brown2020smooth}, face recognition~\cite{kemelmacher2016megaface,maze2018iarpa}, and vehicle re-identification~\cite{khan2019survey,he2019part}.

\begin{figure}[t]
\centering
\includegraphics[width=0.95\linewidth]{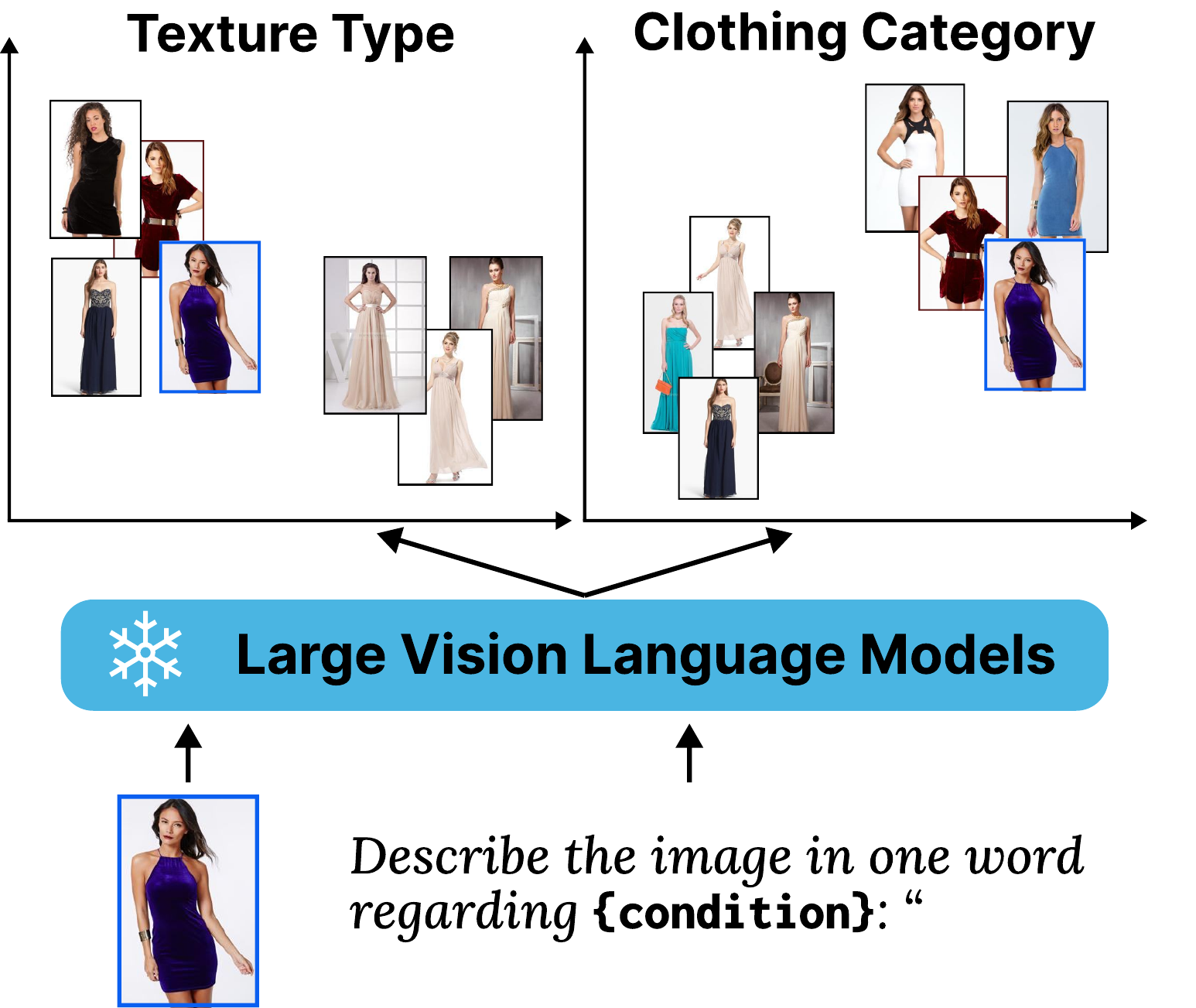}
\caption{Images are projected into vector spaces based on given conditions, such as texture types or clothing categories.
Our proposed method, DIOR, generates conditional embeddings by inputting the image and condition into a large vision-language model, without the need for additional training.}
\label{fig:overview}
\end{figure}

However, relying solely on a single global image embedding has a critical limitation: it often fails to capture specific user intent by blending distinct aspects of an image into a single representation.
As illustrated in \cref{fig:overview}, users might wish to search based on specific attributes, such as clothing textures (e.g., silk, denim) or clothing categories (e.g., dresses, jackets).
Global embeddings can conflate these different search intents, resulting in ambiguity. This ambiguity arises because global embeddings aggregate all visual features, making it challenging to distinguish between distinct criteria.

To address this limitation, the concept of \textit{conditional image similarity}~\cite{veit2017conditional} has emerged. This approach enables precise comparisons by evaluating image similarity based on specified textual conditions.
Such fine-grained control significantly enhances practical applications, allowing more intuitive and targeted interaction with visual data.
In e-commerce, for instance, users can search specifically for clothing items with textures similar to a reference image, regardless of differences in clothing categories, thus substantially improving the product discovery experience.

In this scenario, the embedding used for computing conditional image similarity is referred to as a \textbf{conditional image embedding}.
Early methods relied on supervised learning with predefined conditions~\cite{veit2017conditional,vasileva2018learning,tan2019learning}. 
Visual foundation models like CLIP~\cite{radford2021learning} offer general representations but lack explicit conditioning mechanisms.
To address this issue, a recent study~\cite{kobs2023indirect} proposed deriving conditional image embeddings by linearly mapping CLIP embeddings into condition-specific subspaces.
However, CLIP embeddings often struggle to effectively separate specific image attributes, limiting the utility of these subspaces~\cite{yi2024leveraging,ramasinghe2024accept,huang2024troika}. Moreover, this approach still depends on training with manually curated, condition-related terms.

In this paper, we propose \textbf{DIOR} (\textbf{D}escribe the \textbf{I}mage in \textbf{O}ne word \textbf{R}egarding a given condition), a novel training-free approach to obtain conditional image embeddings by leveraging large vision-language models (LVLMs).
As shown in \cref{fig:overview}, given an image and a text condition, DIOR feeds the image and a prompt ``\texttt{Describe the image in one word regarding \{condition\}:}'' to an LVLM.
This prompting strategy enables the LVLM to summarize the image representation into the hidden state vector of the \red{last prompt token} at the last layer, which serves as the conditional image embedding.
Unlike conventional approaches that primarily use LVLMs for generation tasks, DIOR innovatively repurposes a frozen LVLM as an embedder for conditional image retrieval.
This adaptation leverages the powerful latent representations of LVLMs, enabling flexible and effective conditional image similarity computation without requiring task-specific priors or additional training.

We evaluate DIOR on three benchmarks for conditional image similarity: LanZ-DML~\cite{kobs2023indirect}, Style Similarity~\cite{somepalli2024measuring}, and GeneCIS~\cite{vaze2023genecis}.
DIOR outperforms task-agnostic baselines in most cases and shows promising performance compared to task-specific baselines.
Ablation studies investigate DIOR's design space to underscore critical decisions and demonstrate that certain naive baselines are significantly suboptimal.
By visualizing the embeddings and qualitatively analyzing examples, we demonstrate that DIOR can generate image embeddings conditioned on given attributes.

Our main contributions are as follows:
\begin{itemize}
\item We propose DIOR, a novel conditional image embedding framework. This is the first framework to leverage the visual knowledge and instruction-following capabilities of LVLMs for this task.

\item DIOR does not require additional training and can handle various conditions within a unified framework.

\item We empirically show that DIOR, despite being training-free, achieves competitive or superior performance compared to state-of-the-art models that are specifically trained for each task.
\end{itemize}

%% file: sec/2_related_work.tex
\section{Related Work}
\label{sec:related_work}

\subsection{Image Embeddings}
Traditionally, image embedding methods have focused on building models specialized for each domain.
For example, a variety of models have been built for image retrieval in the food~\cite{min2023large}, products~\cite{oh2016deep}, bird~\cite{WahCUB_200_2011}, car~\cite{krause20133d}, fashion~\cite{liu2016deepfashion}, and landmark~\cite{radenovic2018revisiting,weyand2020google} domains, as well as for person recognition from person-related data such as faces~\cite{kemelmacher2016megaface,maze2018iarpa} or full bodies~\cite{li2014deepreid,zheng2015scalable}.
These models employ convolutional neural networks as backbones and are trained via deep metric learning using embedding and classification losses~\cite{musgrave2020metric,chen2022deep}.
The cost of training and maintaining an embedder model for each specific domain has been mitigated by the development of recent visual foundation models such as CLIP~\cite{radford2021learning} and DINO~\cite{caron2021emerging}.
These models can accurately embed any image from any domain, owing to pre-training on large-scale data either through multi-domain unsupervised learning~\cite{almazan2022granularity} or classification on vast image-label pairs~\cite{ypsilantis2023towards}. Nonetheless, these embeddings, referred to as universal image embeddings, encapsulate the entirety of concepts within an image and are thus unsuitable for conditional image embeddings.

\subsection{Conditional Image Embeddings}
Conditional image embedding aims to generate image embeddings that focus on a specific concept within an image using textual input, allowing similarity to be calculated from the user's desired perspective. The similarity between two such embeddings is termed conditional image similarity~\cite{veit2017conditional}.
Some methods have explored learning subspace embeddings to capture different notions of similarity for fashion~\cite{veit2017conditional,vasileva2018learning,tan2019learning,lin2020fashion}, birds~\cite{mishra2021effectively}, etc. However, these approaches require large-scale datasets of images and a predefined set of attributes associated with each image for a specific domain, and they do not generalize to unseen image domains or attributes.

Since the emergence of vision foundation models, several approaches have been proposed to obtain conditional image embeddings with minimal supervision.
Vaze \etal~\cite{vaze2023genecis} propose to generate image embeddings based on a free-form text condition;
however, their method still relies on training with a large-scale dataset. 
In contrast, our DIOR is training-free yet outperforms their method, as we demonstrate in \cref{sec:result}.

The work closest to ours is an approach by Kobs \etal~\cite{kobs2023indirect}. They generate conditional embeddings by performing Principal Component Analysis (PCA) on CLIP image embeddings. By using a few CLIP text embeddings of related words manually listed for the target condition, they attempt to identify a subspace representing the condition in CLIP features at test time.
However, embeddings from the pre-trained CLIP model do not guarantee a disentangled representation of specific aspects in the latent space~\cite{you2024calibrating,huang2024troika}, which affects similarity calculation between images when used as-is~\cite{yi2024leveraging,ramasinghe2024accept}.
Such embeddings are difficult to decompose by simple PCA.
We argue that DIOR discovers a complex non-linear mapping between universal and conditional image embeddings by leveraging powerful language models within LVLMs, since LVLMs typically rely on recent vision foundation models that produce universal image embeddings, such as CLIP.

\begin{figure}[t]
\centering
\includegraphics[width=\linewidth]{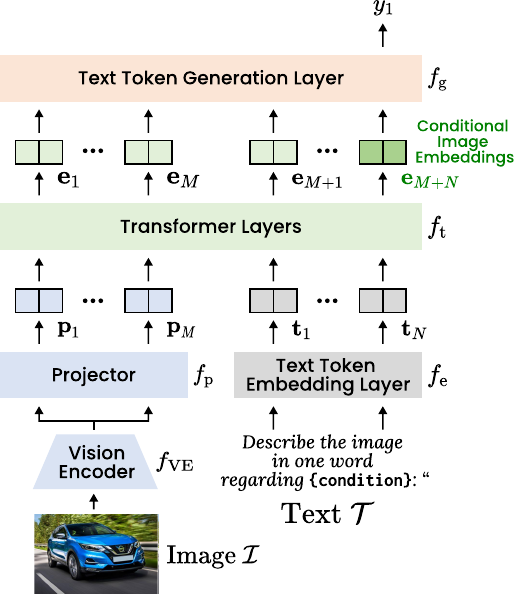}
\caption{Overview of the DIOR framework. 
In our method, we first input the image along with a prompt describing the condition into a large vision-language model.
The output from the final Transformer layer of the final token in the prompt is then used as the conditional image embedding.}
\label{fig:method}
\end{figure}

\subsection{Large Vision-Language Models (LVLMs)}
An LVLM integrates a pre-trained vision encoder, such as a Vision Transformer (ViT)~\cite{dosovitskiy2020image} based on CLIP~\cite{radford2021learning} or BLIP~\cite{li2022blip}, with a pre-trained large language model such as Vicuna~\cite{vicuna2023} or LLaMA~\cite{touvron2023llama}. This integration is accomplished in two phases. First, a projector is trained to align image features with text token features, thereby connecting encoders of two different modalities. Second, the model undergoes visual instruction tuning, enabling it to handle tasks involving both visual and textual input.
Through these phases, the LVLM learns the correspondence between vision and text, enabling it to be applied to a variety of text generation tasks related to images, such as image captioning and visual question answering, without requiring task-specific training~\cite{liu2024visual}. 
Our DIOR framework is versatile and can be applied to any LVLM in a plug-and-play manner.

%% file: sec/3_proposed_method.tex
\section{Proposed Method}
\label{sec:proposed_method}
We propose DIOR, a novel framework for generating conditional image embeddings from a pre-trained large vision-language model (LVLM) without additional training.
We first describe the LVLM architecture and then introduce the proposed DIOR framework.

\subsection{Large Vision-Language Model}
\label{sec:lvlm}
An LVLM connects a pre-trained vision encoder with a pre-trained large language model.
We briefly describe the structure of a typical LVLM as shown in \cref{fig:method}.
Let the input image be denoted as \( \mathcal{I} \), and the text prompt be denoted as \( \mathcal{T} = \{ t_1, \ldots, t_N\} \) with the number of tokens as \( N \). 
The image \( \mathcal{I} \) is first input to a vision encoder \( f_{\rm{VE}} \), and then image features compatible with text token embeddings are generated through a projector \( f_{\rm{p}} \).
The process can be described as:
\begin{equation}
\mathbf{p} = f_{\rm p}(f_{\rm{VE}}(\mathbf{\mathcal{I}})),
\end{equation}
where \( \mathbf{p} = \{ \mathbf{p}_1, \dots, \mathbf{p}_M\} \) represents the projected embeddings of the image and \( M \) is the number of patches produced by the vision encoder.
The text \( \mathcal{T} \) is also converted to token features through a text token embedding layer:
\begin{equation}
\mathbf{t} = f_{\rm{e}}(\mathbf{\mathcal{T}}),
\end{equation}
where \( \mathbf{t} = \{ \mathbf{t}_1, \dots, \mathbf{t}_N\} \) represents the text token embeddings.

The projected embeddings \( \mathbf{p} \) and the text token embeddings \( \mathbf{t} \) are combined with positional embeddings and then input into the large language model (LLM).
The LLM typically consists of a decoder-based Transformer, where \( f_{\rm{t}} \) represents the Transformer layer.
The combined input is processed as:
\begin{equation}
\mathbf{e} = f_{\rm{t}}(\mathbf{p}, \mathbf{t}),
\end{equation}
where \( \mathbf{e} = \{ \mathbf{e}_1, \dots, \mathbf{e}_{M+N}\} \) denotes the combined intermediate representations.
Finally, these intermediate representations \( \mathbf{e} \) are passed through the text token generation layer \( f_{\rm{g}} \), which outputs a text token:
\begin{equation}
y_1 = f_{\rm{g}}(\mathbf{e}_{M+N}),
\end{equation}
where $y_1$ denotes the first output text token\footnote{\red{We adopt the no-shift convention, where the first output token \( y_1 \) is predicted from the hidden state of the last prompt token \( e_{M+N} \).}}.
A new text can be generated by concatenating the output text token to the input text and repeating the same inference procedure.

\subsection{DIOR Framework}
While LVLMs are primarily used to generate text about images, our goal is to generate image embeddings by using the model's visual knowledge and instruction-following capabilities.
In the natural language processing field, similar methods have been proposed that input a prompt into an LLM and use it as a text embedder~\cite{jiang2023scaling,zhang2024simple,yamada2025out}.
The prompt is ``\texttt{This sentence: \{text\} means in one word:}'', where \texttt{\{text\}} is a placeholder for the input text. This prompt is input into the LLM, and the hidden state vector of the \red{last prompt token} is used as the text embedding.
The one-word constraint encourages the model to compress the meaning of the entire text into the hidden state vector of the last token.

We propose an extension of this approach, DIOR, which generates image embeddings that focus on specific aspects of the input image based on a given condition.
An overview of the DIOR framework is shown in \cref{fig:method}. 
We realize conditional image embeddings by using a prompt with a one-word constraint and a specified condition, such as ``\texttt{Describe the image in one word regarding \{condition\}:}''.
By forcing the LVLM to represent the image in a single word while specifying the aspect as the condition, it is expected to compress the visual information in a targeted manner.

The inference process for embedding extraction is no different from the text generation of LVLMs described in \cref{sec:lvlm}.
We feed \( M \) projected embeddings for images and \( N \) text token embeddings for text prompts into the Transformer layer and use the output representation \( \mathbf{e}_{M+N} \) of the last token of the prompt as the conditional image embedding.
Notably, our approach requires no additional training, using the LVLM's inherent understanding capabilities.
This enables the method to be applied effectively across various domains, highlighting its flexibility and broad applicability.

%% file: sec/4_experimental_settings.tex
\section{Experimental Setup}
\label{sec:experiments}
We demonstrate the effectiveness of DIOR on multiple benchmarks for conditional image similarity tasks.
First, we introduce the common experimental settings shared across benchmarks, including an overview of each task and the implementation details for both standard baselines and DIOR.
Then, we provide comprehensive results and ablation studies for DIOR using the LanZ-DML benchmark~\cite{kobs2023indirect}.
Finally, we present additional evaluations on other conditional image similarity benchmarks, such as style similarity~\cite{somepalli2024measuring} and GeneCIS~\cite{vaze2023genecis}\footnote{Detailed experimental settings and results for GeneCIS are provided in the Supplementary Material~\ref{sec:genecis}.}, highlighting the broader applicability of DIOR.

\subsection{Common Settings and Our Proposed Method}
\paragraph{Conditional Image Similarity Task.}
In this task, we retrieve the target image similar to the query image from an indexed set of images based on specific textual conditions.
The conditions include, for example, the object's name, color, or painting style.
Conditional image similarity is measured as follows.
Let \( \mathcal{I}^{\rm{query}} \) be the query image and \( \mathcal{I}^{\rm{index}} \) be the index image.
We use the image embeddings \( \mathbf{e}_c^{\rm{query}} \) and \( \mathbf{e}_c^{\rm{index}} \) for \( \mathcal{I}^{\rm{query}} \) and \( \mathcal{I}^{\rm{index}} \) under condition \(c\), respectively. 
\red{
The similarity between image embeddings is computed using a similarity function \( f_{\rm{sim}} \), such as cosine similarity:
\(
f_{\rm{sim}}(\mathbf{e}_c^{\rm{query}}, \mathbf{e}_c^{\rm{index}})
\) which we used in all experimental settings.}

\paragraph{Common Baseline Method.}
We use CLIP~\cite{radford2021learning}, which has been trained using contrastive learning~\cite{oord2018representation} on a large dataset of image-text pairs to associate visual content with textual descriptions, as the image embedder.
There are several variations based on the size of the model or the image patches input to the Vision Transformer (ViT)~\cite{dosovitskiy2020image}. 
We use CLIP ViT-L/14 as the base model.

\input{tables/result_lanz_dml_modified}

\paragraph{Our Proposed Method.}
DIOR uses LVLMs to create conditional image embeddings as described in~\cref{sec:proposed_method}.
We employ Llama-3.2-Vision-Instruct (11B)\footnote{\href{meta-llama/Llama-3.2-11B-Vision-Instruct}{meta-llama/Llama-3.2-11B-Vision-Instruct}}\cite{meta2024llama} as our primary backbone LVLM, while also utilizing LLaVA-1.6(7B/13B)\footnote{\href{https://huggingface.co/llava-hf/llava-v1.6-vicuna-7b-hf}{llava-v1.6-vicuna-7b-hf} and \href{https://huggingface.co/llava-hf/llava-v1.6-vicuna-13b-hf}{llava-v1.6-vicuna-13b-hf}}~\cite{liu2024visual}.
For a fair evaluation under conditions where the vision encoder is identical to that of the baseline CLIP, we selected these three models that use CLIP ViT-L/14 as their vision encoder.
Furthermore, to verify the applicability of our proposed method to newly released models, we include the recently released high-performance Qwen-2.5-VL (3B/7B)\cite{bai2025qwen25vltechnicalreport}\footnote{\href{https://huggingface.co/Qwen/Qwen2.5-VL-3B-Instruct}{Qwen2.5-VL-3B-Instruct} and \href{https://huggingface.co/Qwen/Qwen2.5-VL-7B-Instruct}{Qwen2.5-VL-7B-Instruct}} in our evaluations.
Unless otherwise specified, we use the default prompt template:
``Describe the image in one word regarding {condition}:''
which enforces both a one-word and conditional constraint.

\subsection{Benchmark-Specific Settings and Baselines}

\paragraph{LanZ-DML.}

LanZ-DML~\cite{kobs2023indirect} is a benchmark to verify the ability to generate image representations for each different aspect, such as color and category, assigned to the same image. 
Following~\cite{kobs2023indirect}, we experiment on five datasets: Synthetic Cars~\cite{kobs2021different}, Cars196~\cite{krause20133d}, CUB200~\cite{WahCUB_200_2011}, DeepFashion~\cite{liu2016deepfashion}, and Movie Poster~\cite{chu2017movie}.
The conditions are ``Car Model'', ``Car Color'', and ``Background Color'' for Synthetic Cars, a ``Car Model'' for Cars196, a ``Bird Species'' for CUB200, a ``Clothing Category'', ``Texture'', ``Fabric'', and ``Fit'' for DeepFashion, and a ``Genre'' and ``Production Country'' for Movie Poster.
We retrieve the image from the index images that have the same features for the query image under the conditions.
Following prior settings, we report Mean Average Precision at R (MAP@R) as evaluation metrics.
As an additional baseline specific to this benchmark, we use InDiReCT~\cite{kobs2023indirect}.
While the original implementation used given specific candidates related to the condition, we also provide an implementation that automatically generates candidates with GPT-4o and uses them for PCA to make a more realistic and fair setting that matches the settings of our proposed method.
We refer to the original and the fairer settings as InDiReCT~(Oracle) and InDiReCT~(Real), respectively, and report both results.
\red{To evaluate the effects of scaling up vision encoder-only models such as CLIP, we also conduct experiments using EVA-CLIP~(18B)~\cite{evaclip}. Specifically, we perform comparisons not only using embeddings directly extracted from the vision encoder but also by using EVA-CLIP~(18B) as a backbone model within InDiReCT.}
We also introduce another naive baseline, CapEmb. CapEmb is a two-stage approach designed to simulate the naive adaptation of LVLMs for the task unlike DIOR. First, it generates text descriptions using the same prompt as DIOR for an LVLM (i.e., Llama-3.2 11B). Second, these texts are converted into embeddings using a text embedder model. We employ two widely recognized text embedders: SimCSE~\cite{gao-etal-2021-simcse} and SentenceT5~\cite{ni2021sentence}~\footnote{We also tested a CapsEmb baseline that performed captioning without the single-word constraint in prompts. However, since performance improved with the single-word constraint, we adopted the baseline that included it.}.

\paragraph{Style Similarity.}

We examine DIOR on a more challenging task, measuring image style similarity~\cite{somepalli2024measuring}.
We employ two datasets WikiArt~\cite{saleh2015large} and DomainNet~\cite{peng2019moment}.
WikiArt is a dataset focused on image styles, including various artistic styles.
DomainNet is a dataset that includes images from various domains and features two conditions: `Style' and `Object'. 
While the previous study focused solely on style, we conduct experiments that also consider the object. Since the dataset split from the prior study is not available, we create a new subset for this experiment.
The evaluation metrics are Mean Average Precision at k~(MAP@k) and Recall at k~(Recall@k), with evaluations conducted for k=1, 10, and 100.
As an additional baseline for Style Similarity tasks, we use the CLIP baselines and CSD~\cite{somepalli2024measuring}, a model that fine-tunes CLIP L/14 using contrastive learning on style-labeled images.

\input{tables/results_domainnet}
\input{tables/results_wikiart}

\section{Results and Analysis}
\subsection{Main Results}
\label{sec:result}

We first present results from our primary benchmark, LanZ-DML, to demonstrate the fundamental effectiveness of DIOR compared to various baselines. As shown in \cref{tab:result_lanz_dml_transposed}, among models sharing the same vision encoder architecture (Llama-3.2, LLaVA-1.6-13B, LLaVA-1.6-7B, and CLIP), DIOR with Llama-3.2 exhibits the highest overall performance, consistently outperforming all other baselines.This indicates that DIOR effectively extracts image features conditioned on textual input.
Additionally, Qwen-2.5-VL, a relatively recent and smaller model, achieves a higher average score than Llama-3.2, indicating that DIOR directly benefits from ongoing advancements in LVLMs. This performance trend aligns with the general behavior of Vision-Language Models (VLMs) reported in prior studies~\cite{duan2024vlmevalkit}, suggesting that employing advanced VLMs facilitates the generation of more robust and higher-quality conditional image embeddings. Notably, CapEmb's performance is sometimes even inferior to CLIP's, emphasizing the importance of our approach of directly obtaining embeddings from the hidden states of LVLMs without decoding tokens.
\red{Examining the results from InDireCT using EVA-CLIP~(18B) as a backbone~InDireCT~(EVA-CLIP), we find that it surpasses DIOR in certain conditions such as Synthetic Cars' Model and CUB200. However, on datasets with multiple conditions, including Synthetic Cars, DeepFashion, and Movie Poster, InDireCT~(EVA-CLIP) fails to adequately handle changes in the relevant conditions. This observation indicates that merely scaling up the parameter size of vision encoder-only models like CLIP does not sufficiently address the conditional image similarity task.}

To evaluate its generalization capabilities, we test DIOR on the more challenging task of style similarity.
\cref{tab:result_domainnet_newcol,tab:result_wikiart_newcol} show the results for DomainNet and WikiArt, respectively.
On DomainNet, DIOR surpasses other approaches in both style and object similarity.
While CSD enhances performance in style similarity through style-specific fine-tuning, it compromises object similarity. 
In contrast, DIOR's training-free nature enables it to achieve superior performance in both similarity metrics simultaneously.
On the other hand, for the highly specific domain of WikiArt, we observed that DIOR performed below the baseline methods, CLIP and CSD. We will discuss this limitation in Supplementary Material~\ref{sec:peformance_degration}.

\begin{figure}[t]
\centering
\includegraphics[width=\linewidth]{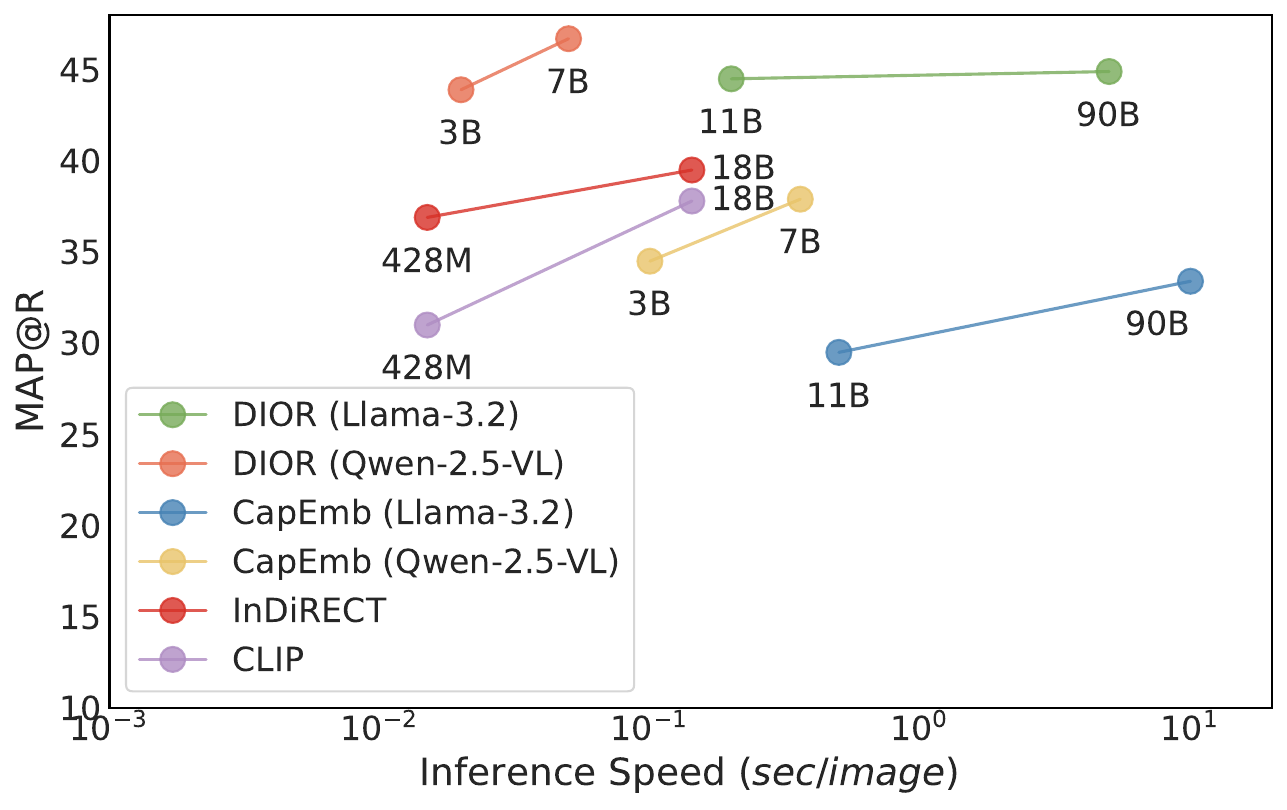}
\caption{Relationship between inference speed and average performance for each model on the LanZ-DML Benchmark.
We use SentenceT5 as the text embedder for CapEmb.
\red{In InDiRECT and CLIP, ``428M'' and ``18B'' refer to CLIP L/14 and EVA-CLIP~(18B), respectively. }
}
\label{fig:inference_speed}
\end{figure}

\input{tables/results_cache}

\subsection{Analysis}
\label{sec:analysis}

\paragraph{Speed-Performance}
\red{To illustrate the trade-off between performance and inference speed, we plotted the average score on the LanZ-DML benchmark against the inference time per image, as shown in \cref{fig:inference_speed}. The results reveal that our proposed method, DIOR, generally outperforms the baselines in accuracy but incurs a slower inference speed due to its larger LVLM architecture.
While simply scaling up a vision-encoder-only model like EVA-CLIP~(18B) improves performance over the standard CLIP baseline, our DIOR framework using more recent and efficient LVLMs achieves superior results. For instance, DIOR with 
Qwen-2.5-VL as its backbone achieves high performance and faster inference despite its smaller parameter size. This indicates that the advanced reasoning capabilities of LVLMs are more beneficial than merely increasing the scale of the vision encoder.
Furthermore, among the LVLM-based approaches, DIOR consistently outperforms the naive 
CapEmb baseline in both performance and speed metrics, highlighting the effectiveness of directly using hidden states for embeddings.
We also evaluated a significantly larger LVLM, llama-3.2-Vision-Instruct (70B), but observed only modest gains in MAP@R scores relative to a substantial reduction in inference speed. 
This result further highlights the efficiency and effectiveness of newer, optimized models.
}

Our method also incurs higher computational costs compared to vision-encoder-only approaches, such as CLIP. In practical scenarios, this becomes notably inefficient when multiple embeddings must be generated from a single image under varying conditions. To address this inefficiency, we use KV-caching to minimize redundant computations. Specifically, by caching outputs from the image encoder and the static portion of the language prompt ("\texttt{Describe the image in one word regarding}"), our method executes only the essential computations required by each specific condition.
This caching approach significantly reduces computational overhead, achieving roughly a 55\% decrease compared to inference without caching, as demonstrated in \cref{tab:cache_speed}.

\begin{figure}[t]
\centering
\includegraphics[width=\linewidth]{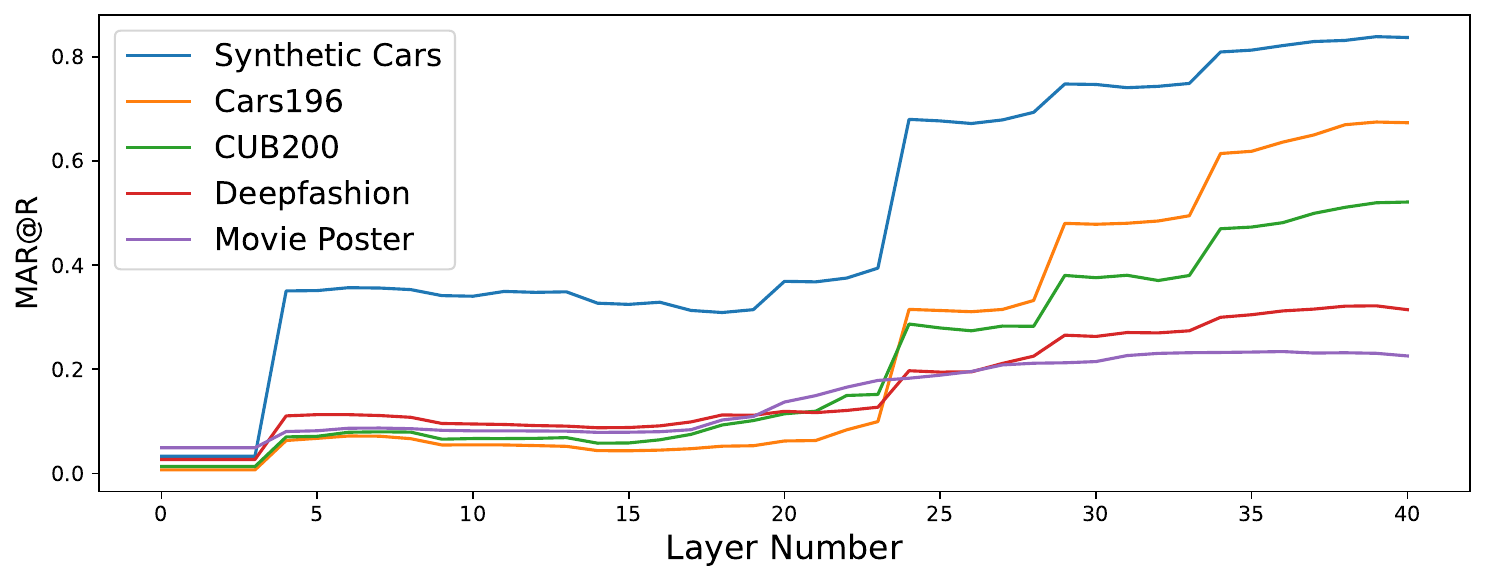}
\caption{Comparison of MAP@R for each layer's hidden state.
}
\label{fig:layer_analysis}
\end{figure}

\input{tables/results_compare_token}

\begin{figure*}[t]
\small
\centering
\includegraphics[width=0.9\linewidth]{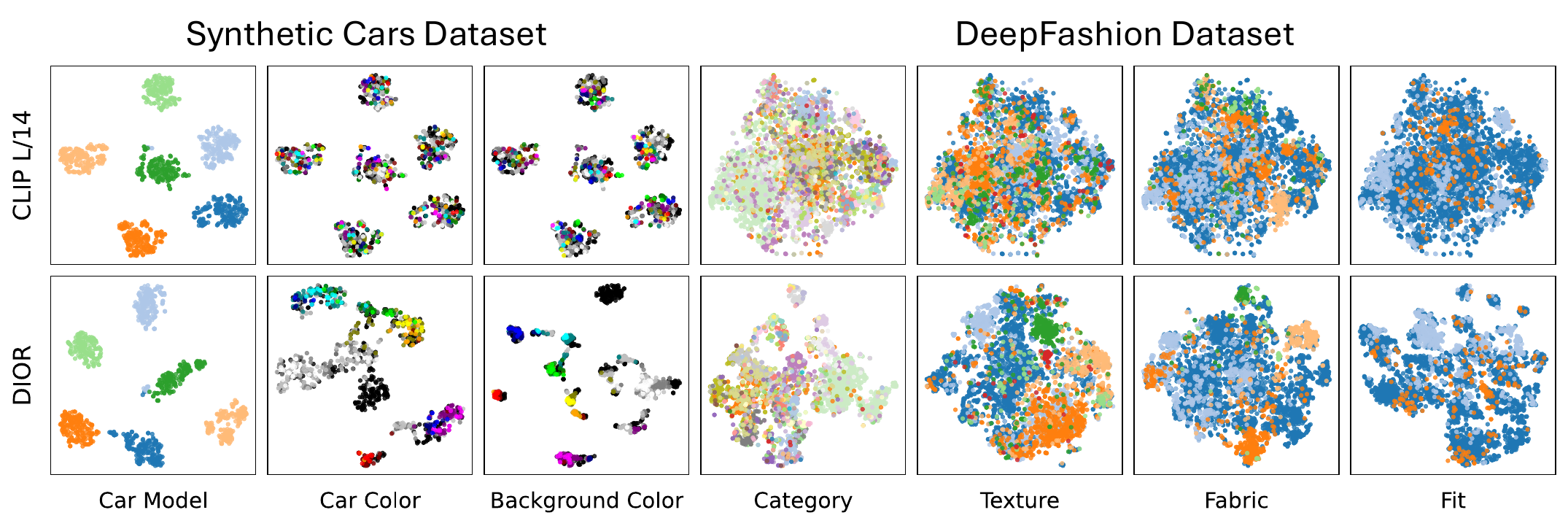}
\caption{2D t-SNE projections of the image embeddings by CLIP L/14 and DIOR on the Synthetic Cars and DeepFashion datasets.
Each color indicates the class type in each condition.}
\label{fig:compare_embedding}
\end{figure*}

\input{tables/result_prompt_type}

\paragraph{Validity of Embedding Extraction Method in DIOR}
\red{Regarding the validity of the embedding extraction method in DIOR, embeddings are obtained from the last token of the prompt at the final transformer layer of the LVLM. However, several alternative extraction methods, particularly comparing embeddings from different layers and tokens, can also be considered.}

For layer selection, DIOR has numerous options due to the multiple transformer blocks within base LVLMs, as illustrated in \cref{fig:method}. We analyze DIOR's performance across different layers. As shown in \cref{fig:layer_analysis}, layers closer to the final text generation layer $f_{\rm{g}}$ yield superior embeddings, with the optimal representation obtained from the layer immediately preceding the final text generation layer.
\red{
Regarding token selection, we compare three approaches in \cref{tab:results_compare_token}: the final token of the input prompt (last input), the first generated token (first output), and the average of generated tokens (mean output). The first output and mean output embeddings are acquired from greedy decoding (temperature=0) at the final layer. Results show that the last input embeddings slightly outperform or match the first output, while mean output embeddings consistently perform worse. Given that the first output embeddings can be influenced by decoding temperature and involve extra computational cost, using the last input token for embedding extraction is recommended.
}

\input{tables/results_ami}

\paragraph{Clustering Tasks}
Image embeddings projected into two dimensions by t-SNE~\cite{van2008visualizing} are shown in \cref{fig:compare_embedding}.
\red{For a more quantitative comparison, we also report Adjusted Mutual Information~(AMI) in \cref{tab:ami_scores}.
This plot compares embeddings generated by CLIP L/14 and DIOR with Llama-3.2 on the Synthetic Car and DeepFashion datasets, which includes multiple conditions for each image. 
AMI scores indicate that DIOR generally achieves better-defined clustering compared to CLIP.
This improvement is particularly evident for attributes such as `Car Color' on Synthetic Cars (0.297 vs. 0.011) and multiple attributes on DeepFashion, including `Clothing Category' (0.429 vs. 0.251), `Texture' (0.164 vs. 0.078), and `Fit' (0.058 vs. 0.002). 
CLIP can cluster primary objects in the image, such as a car model, but its ability to focus on specific areas and capture visual differences seems relatively poor.}

\paragraph{Robustness to Prompt Variation}
We examine the impact of prompt variation on DIOR's performance, as detailed in \cref{tab:result_prompt_type}.
As illustrated in the top half, the method using both one-word and conditional restrictions is the highest performance, showing that both restrictions are critical.

In the bottom half, we test slight prompt variation by changing the verb ``Describe'' to other verbs. The performance remains stable except for a slight decline by ``Depict''.
``Depict'' originally means ``to illustrating'' or ``visually representing'', but it may not align well with the task's requirements .
In short, as long as a single word and conditional restrictions are specified, performance will not be affected.

%% file: tables/result_lanz_dml_modified.tex
\setlength{\tabcolsep}{4.5pt}
\begin{table*}[t]
\centering
\caption{Evaluation results for the LanZ-DML benchmark. Since we could not obtain the text labels from the metadata of Cars196, they are marked as N/A in InDiReCT. 
\red{* indicates that the vision encoder is not CLIP-L/14 and $\dagger$ denotes methods that require training.}
Avg represents the average over the respective conditions for the Synthetic Cars, Cars196~(Cars), CUB200~(CUB), DeepFashion, and Movie Posters datasets. Since InDiReCT's performance varies significantly depending on the number and types of text labels used, we prepared a separate development subset (dev) apart from the test data. We then selected the label set that achieved the highest performance on the dev subset and used it to evaluate on the test data. See Supplementary Material~\ref{lanz-dml-experiments} for details.}
\begin{tabular}{@{}lrcccccccccccc@{}}
\toprule
\multicolumn{1}{c}{} & \multicolumn{1}{c}{}
& \multicolumn{3}{c}{Synthetic Cars}
& \multicolumn{1}{c}{Cars}
& \multicolumn{1}{c}{CUB}
& \multicolumn{4}{c}{DeepFashion}
& \multicolumn{2}{c}{Movie Posters} &  \\

\cmidrule(lr){3-5}\cmidrule(lr){6-6}\cmidrule(lr){7-7}\cmidrule(lr){8-11}\cmidrule(lr){12-13}
& Params
& Mod. & Col. & BG
& Mod.
& Spe.
& Clo. & Tex. & Fab. & Fit
& Gen. & Cou. & Avg\\
\midrule
CLIP                 &  428M  & 72.3 & 5.8 & 6.8 & 43.0 & 33.3 & 12.1 & 18.4 & 33.9 & 53.0 & 12.2 & 50.7 & 31.0 \\
EVA-CLIP*            & 18B & 91.9 & 5.9 & 6.8 & 67.7 & 60.2 & 15.3 & 18.8 & 35.9 & 53.5 & 10.1 & 49.7 & 37.8 \\
\midrule

InDiReCT~(CLIP)$^\dagger$    &  428M  & 86.3 & 7.6 & 7.9 & N/A & 45.5 & 26.8 & 24.2 & 39.4 & 55.3 & 17.7 & 58.0 & 36.9 \\

InDiReCT~(EVA-CLIP)*$^\dagger$  & 18B & \textbf{94.0} & 7.5 & 7.8 & \text{N/A} & \textbf{66.9} & 32.1 & 24.0 & 40.6 & 54.9 & 14.6 & 52.3 & 39.5 \\

\midrule
CapEmb~(SimCSE)          & 11B   & 51.2 & 11.5 & 11.5 & 28.6 & 24.4 & 23.0 & 13.9 & 39.4 & 52.4 & 18.3 & 53.6 & 29.8 \\
CapEmb~(SentenceT5)      & 11B   & 49.6 & 14.8 & 12.7 & 24.8 & 23.1 & 25.7 & 14.6 & 39.2 & 52.4 & 17.9 & 49.9 & 29.5 \\
\midrule
DIOR~(LLaVA-1.6)      & 7B  & 73.7 & 13.9 & 15.6 & 40.6 & 21.7 & 26.7 & 26.6 & 42.8 & 54.5 & 20.3 & 58.1 & 35.9 \\
DIOR~(LLaVA-1.6)     & 13B & 78.9 & 14.9 & 15.4 & 45.8 & 23.1 & 30.8 & \textbf{34.8} & 43.4 & 57.4 & 21.9 & 60.0 & 38.8 \\
DIOR~(Llama-3.2)     & 11B & 88.0 & 14.7 & \textbf{17.8} & 67.3 & 51.9 & 33.7 & 26.0 & 46.4 & 57.8 & 23.4 & 62.2 & 44.5 \\
DIOR~(Qwen-2.5-VL)* & 3B  & 71.0  & 14.4 & 15.3  & 65.5 & 57.7  & 42.7  & 31.2 & 43.6 & 58.4  &  21.5 & 61.1 & 43.9 \\
DIOR~(Qwen-2.5-VL)* & 7B  & 73.2  & \textbf{15.0} & 16.8  & \textbf{68.0} & 58.1 & \textbf{45.3} &  33.9  & \textbf{47.6}   &  \textbf{62.1} & \textbf{28.1}  & \textbf{66.0} & \textbf{46.7} \\
\bottomrule
\end{tabular}
\label{tab:result_lanz_dml_transposed}
\end{table*}

%% file: tables/results_domainnet.tex

\begin{table*}[t]
\centering
\caption{Evaluation Results for DomainNet.}
\begin{tabular}{@{}lccccccccccccc@{}}
\toprule
& 
& \multicolumn{3}{c}{Style (MAP@k)} 
& \multicolumn{3}{c}{Style (Recall@k)} 
& \multicolumn{3}{c}{Object (MAP@k)} 
& \multicolumn{3}{c}{Object (Recall@k)} \\
\cmidrule(lr){3-5} \cmidrule(lr){6-8} \cmidrule(lr){9-11} \cmidrule(lr){12-14}
Method & Train
& 1 & 10 & 100 
& 1 & 10 & 100 
& 1 & 10 & 100 
& 1 & 10 & 100 \\
\midrule
CLIP ViT-L/14  & 
& 81.74 & 77.86 & 69.20 
& 81.74 & \textbf{97.30} & \textbf{99.87} 
& 69.21 & 64.13 & 50.86 
& 69.21 & 87.78 & \textbf{96.25}  \\

CSD ViT-L/14 & \checkmark
& 83.50 & 80.41 & 75.09 
& 83.50 & 97.08 & 99.89
& 61.27 & 54.11 & 37.24 
& 61.27 & 85.03 & 95.37  \\

DIOR & 
& \textbf{83.57} & \textbf{81.10} & \textbf{76.81}
& \textbf{83.57} & 97.17 & 99.80
& \textbf{72.55} & \textbf{68.27} & \textbf{57.62}
& \textbf{72.55} & \textbf{88.86} & 96.06  \\
\bottomrule
\end{tabular}
\label{tab:result_domainnet_newcol}
\end{table*}

%% file: tables/results_wikiart.tex

\begin{table}[t]
\centering
\caption{Evaluation Results for Wikiart.}
\setlength{\tabcolsep}{3pt}
\begin{tabular}{@{}lccccccc@{}}
\toprule
& & \multicolumn{3}{c}{MAP@k} & \multicolumn{3}{c}{Recall@k} \\
\cmidrule(lr){3-5} \cmidrule(lr){6-8}
Method & Train & 1 & 10 & 100 & 1 & 10 & 100 \\
\midrule
CLIP ViT-L/14  &
& \textbf{59.3} & \textbf{48.7} & \textbf{31.5}
& \textbf{59.3} & \textbf{82.9} & \textbf{95.1} \\

CSD ViT-L/14      &  \checkmark
& 58.2 & 48.0 & 30.8
& 58.2 & 81.9 & 94.5 \\

DIOR           & 
& 45.6 & 39.1 & 28.9
& 45.6 & 71.0 & 89.0 \\
\bottomrule
\end{tabular}
\label{tab:result_wikiart_newcol}
\end{table}

%% file: tables/results_cache.tex
\begin{table}[t]
    \caption{Comparison of total inference times (ms) for three inferences across three conditions, evaluated on the Synthetic Cars dataset.}
    \centering
    \label{tab:inference_times}
    \begin{tabular}{@{}lcc@{}}
        \toprule
        \textbf{Method} & \textbf{w/o cache} & \textbf{w/ cache} \\
        \midrule
        CLIP L/14 & 36.10 & -- \\
        \midrule
        Llama-3.2~(11B)  & 549.72 &  244.40~($\downarrow$ 55.55\%)\\
        Qwen-2.5-VL~(3B) & 200.40 & {\space\space 89.90}~($\downarrow$ 55.14\%) \\
        \bottomrule
    \end{tabular}
    \label{tab:cache_speed}
\end{table}

%% file: tables/results_compare_token.tex


\begin{table}[t]
  \centering
  \caption{Comparison of the average MAR@k scores for tokens used as conditional embeddings in DIOR. SC, Car, CUB, DF, and MP denote Synthetic Cars, Cars196, CUB200, DeepFashion, and Movie Poster, respectively.}
  \label{tab:results_compare_token}
  \begin{tabular}{@{}lcccccc@{}}
    \toprule
    & SC & Cars & CUB & DF & MP & Avg \\
    \midrule
    last input   & \textbf{40.2} & 67.3 & \textbf{51.9} & \textbf{41.0} & 42.8 & \textbf{44.5} \\
    first output & 40.1 & \textbf{69.4} & 50.3 & 39.0 & \textbf{45.7} & 44.3 \\
    mean output  & 29.2 & 63.9 & 35.1 & 34.2 & 41.3 & 36.9 \\
    \bottomrule
  \end{tabular}
\end{table}

%% file: tables/result_prompt_type.tex
\begin{table*}[t]
\centering
\setlength{\tabcolsep}{2.5pt}
\caption{Ablation studies on various prompt variations in DIOR. Only MAP@R is reported due to limited space. The bolded value indicates the highest score, and the underlined value indicates the second highest score.}
\begin{tabular}{@{}lccccccc@{}}
\toprule
\multirow{2}{*}{Prompt Type} & \multicolumn{3}{c}{Synthetic Cars} & \multicolumn{4}{c}{DeepFashion} \\
            & Mod. & Col. & BG & Clo. & Tex.& Fab. & Fit \\
\midrule
\texttt{Describe the image in one word regarding \{condition\}:"}  & \textbf{88.0}  & \underline{14.7} & \textbf{17.8} & \underline{33.7} & \textbf{26.0}  & \textbf{46.4} & \textbf{57.8} \\
\texttt{Describe the image in one word:" } & 76.6  & {6.6} & { 5.6} & 17.8 & 16.6  & 34.9  & 53.6  \\
\texttt{Describe the image regarding \{condition\}:" } & 56.5 & { 7.8} & { 8.9} & 12.6 &  14.4 &  37.1  &  55.4 \\
\midrule
\texttt{Express the image in one word in terms of \{condition\}:"}  & \underline{87.9} & \underline{14.7} & \underline{17.5} & \textbf{33.8} & \underline{24.2}  & \underline{45.4}  & 57.4\\
\texttt{Summarize the image in one word regarding \{condition\}:"}  & 84.9 & \textbf{14.8} & 17.6 & 31.7 & 21.5  & 44.5  & \underline{57.7} \\
\texttt{Capture the image in one word based on \{condition\}:"}     &  87.8& 13.8 & 17.0 & 28.3 & 18.4  & 42.6  & 56.9\\
\texttt{Depict the image in one word, considering \{condition\}:" } & 81.3 & \underline{14.7} & 14.2 & 27.9 & 20.4  & 40.7  & 57.0  \\

\bottomrule
\end{tabular}
\label{tab:result_prompt_type}
\end{table*}

%% file: tables/results_ami.tex
\begin{table}[t]
\setlength{\tabcolsep}{2.0pt}
\centering
\caption{AMI~(Adjusted Mutual Information) scores on Synthetic Cars and DeepFashion.}
\label{tab:ami_scores}
\begin{tabular}{l ccc cccc}
\toprule
& \multicolumn{3}{c}{Synthetic Cars} & \multicolumn{4}{c}{DeepFashion} \\
\cmidrule(lr){2-4} \cmidrule(lr){5-8}
\textbf{Method} & Mod. & Col & BG & Clo. & Tex. & Fab. & Fit \\
\midrule
CLIP L/14 & \textbf{0.882} & 0.011 & 0.318 & 0.251 & 0.078 & 0.047 & 0.002 \\
DIOR & 0.823 & \textbf{0.297} & \textbf{0.322} & \textbf{0.429} & \textbf{0.164} & \textbf{0.171} & \textbf{0.058} \\
\bottomrule
\end{tabular}
\end{table}

%% file: sec/5_results.tex
\section{Discussion}
\label{sec:discussion}
Although DIOR is training-free, it requires a little prompt engineering.
We aimed for a minimum and simple prompt in this study, but there may be prompts that perform better without fine-tuning by making use of chain-of-thought prompting~\cite{wei2022chain,kojima2022large} or few-shot prompting~\cite{brown2020language}, which we leave it as a future work.
In addition, image embedding for retrieving images with additional elements given by text information to the image, called compositional image retrieval~\cite{vo2019composing,wu2021fashion,liu2021image}, has not been realized in this study.
This could be solved by improving the inference capability of the model and prompt engineering.
This is also our future work.
DIOR relies on billion-scale LVLMs, which is typically more than ten times larger than the image embedder such as CLIP and thus slow.
We are optimistic since our DIOR can take advantage of rapidly growing smaller but better LVLMs or efficient inference technologies that leverage hardware improvements.

%% file: sec/6_conclusion.tex
\section{Conclusion}

In this paper, we propose a novel framework, DIOR (Describe the Image in One Word Regarding a given condition), to obtain conditional image embeddings.
This method uses an LVLM as its backbone, allowing it to achieve high-quality image embeddings across various domains simply by specifying conditions in the prompt without requiring task-specific training.
In experiments across prior conditional image embedding setups such as LanZ-DML, Style Similarity, and GeneCIS, DIOR demonstrated superior performance over conventional methods.
Qualitative analysis shows that DIOR can generate embeddings emphasizing prompt conditions, though it may be less effective for tasks needing deep expertise in specialized domains.

%% file: sec/7_acknowledgments.tex
\section*{Acknowledgments}
We thank Dr. Kota Yamaguchi for his helpful comments and insightful suggestions.

%% file: sec/X_supplementary.tex
\clearpage
\setcounter{page}{1}
\setcounter{section}{0}
\setcounter{table}{0}
\setcounter{figure}{0}
\setcounter{footnote}{0}

\makeatletter
\@addtoreset{table}{section}
\@addtoreset{figure}{section}
\makeatother
\renewcommand{\thesection}{\Alph{section}}
\renewcommand{\thetable}{\thesection-\arabic{table}} 
\renewcommand{\thefigure}{\thesection-\arabic{figure}}

\maketitlesupplementary

\section{Datasets}

\setlength{\tabcolsep}{5pt}
\begin{table*}[t]
\centering
\caption{Statistics of the datasets used for each task: LanZ-DML and Style Similarity GeneCIS.
* indicates the number of index images per query image.}
\label{tab:sup_datasets}
\begin{tabular}{@{}llcc|cc@{}}
\toprule
\textbf{Task} & \textbf{Dataset} & \textbf{\#Query Image} & \textbf{\#Index Image} & \textbf{Condition} & \textbf{\#Condition Class}\\ \midrule
\multirow{13}{*}{LanZ-DML} & \multirow{3}{*}{Synthetic Cars} & \multirow{3}{*}{1,000} & \multirow{3}{*}{-} & Car Model & 6\\ 
 &  &  &  & Car Color & 18\\ 
 &  &  &  & Background Color &  18\\ \cmidrule{2-6}
 & Cars196 & 8,131  & - & Car Model & 98\\ \cmidrule{2-6}
 & CUB200 & 4,462  & - & Bird Species & 100 \\  \cmidrule{2-6}
 & \multirow{4}{*}{DeepFashion} & \multirow{4}{*}{4,000} & \multirow{4}{*}{-} & Clothing Category & 50\\ 
 &  &  &  & Texture & 7\\ 
 &  &  &  & Fabric & 6\\ 
 &  &  &  & Fit &  3\\ \cmidrule{2-6}
 & \multirow{2}{*}{Movie Posters} & \multirow{2}{*}{8052} & \multirow{2}{*}{-} & Genre & 25\\
 &  &  &  & Production Country & 69\\ 
\midrule
\multirow{3}{*}{Style Similarity}  & Wikiart & 16,006 & 64,090 & Style & 1,119\\ \cmidrule{2-6}
 & \multirow{2}{*}{DomainNet} & \multirow{2}{*}{10,000} & \multirow{2}{*}{100,000} & Object & 345\\
 &  &  &  & Style & 5\\ \midrule
\multirow{2}{*}{GeneCIS} & \multirow{2}{*}{GeneCIS benchmark} & 2,000 & 10*  &  Attribute & 2,000 \\
 &  & 1,960 & 15* &  Object & 1,960 \\
\bottomrule
\end{tabular}
\end{table*}

\cref{tab:sup_datasets} shows the datasets used in our experiments with Language-Guided Zero-Shot Deep Metric Learning (\textbf{LanZ-DML})~\cite{kobs2023indirect}, \textbf{Style Similarity}~\cite{somepalli2024measuring}, and General Conditional Image Similarity (\textbf{GeneCIS})~\cite{vaze2023genecis}.

\paragraph{\textbf{LanZ-DML}}
The LanZ-DML task includes five datasets: Synthetic Cars, Cars196, CUB200, DeepFashion, and the Movie Poster dataset.
The Synthetic Cars dataset consists of artificially generated images of cars on solid-colored backgrounds. The conditions for this dataset include the car model, background color, and car color. Cars196 is also a dataset related to cars, but unlike the Synthetic Cars dataset, it contains real images of cars along with their actual backgrounds.
The Cars196 dataset, unlike the Synthetic Cars Datasets, contains images of real cars.
In previous studies, experiments were conducted using the Cars196 dataset with three conditions: car model, manufacturer, and car type.
However, since the metadata was not available, this experiment was conducted using only the car model as the condition.
CUB200 is a dataset that contains 200 different bird species. The only condition for this dataset is the bird species.
Following the approach of previous studies, 100 species were selected from the 200 available in the dataset for the experiment.
The DeepFashion dataset is a clothing dataset, and it includes four conditions: Clothing Category, Texture, Fabric, and Fit, making it the most diverse in terms of conditions. The Movie Poster dataset consists of movie posters and includes two conditions: movie genre and production country. For all these datasets, including Synthetic Cars, Cars196, CUB200, DeepFashion, and Movie Poster, there is no separate index image, so the query image itself is used as the index image for retrieval.

\paragraph{\textbf{Style Similarity}}
The Style Similarity task includes two datasets: DomainNet and WikiArt.
Wikiart is a dataset focused on image styles and includes a wide variety of artistic styles. The number of different styles in this dataset is 1,119, making it the most diverse among the datasets used. Separate index images are provided, with a total of 16,006 query images and 64,090 index images.
DomainNet is a dataset that includes images from various domains and features two conditions: Style and Object. The Style condition includes six types: Clipart, Infograph, Painting, Quickdraw, Real, and Sketch.
However, since Quickdraw is very similar to Sketch, Quickdraw is excluded from experiments, following the approach of previous studies.
The Object condition consists of 345 categories, including airplane, train, and dog. Domain Net also has index images separate from the query images, with the total number of index images being 238,087.
While previous studies focused solely on style, we conduct experiments that also consider the object.\footnote{Since the dataset split from previous studies was not available, a new subset was created for this experiment, consisting of 10,000 images as query images and 100,000 images as index images. This split will be publicly available on GitHub.}

\paragraph{\textbf{GeneCIS}}
GeneCIS benchmark, unlike other datasets, includes two types of tasks: focus and change. The focus task, similar to other datasets, involves obtaining embeddings that highlight features or objects present in the images.
In contrast, The change task involves creating embeddings for synthesized images that highlight specific features according to given conditions.
Each task—focus and change—includes two conditions: object and attribute. 
In our experiments, we perform evaluations for the focus task under two conditions: attribute and focus.
For each query image in the GeneCIS benchmark, a set of 10 or 15 index images is provided, of which only one is the correct match. 
The evaluation measures the system's ability to accurately retrieve the correct image.
Furthermore, since the GeneCIS dataset includes different conditions for each query image, the number of condition classes corresponds to the number of query images.

\section{LanZ-DML Benchmark Experiments}
\paragraph{\textbf{InDiReCT Baseline}}
\label{lanz-dml-experiments}

\begin{figure*}[t]
\centering
\includegraphics[width=\linewidth]{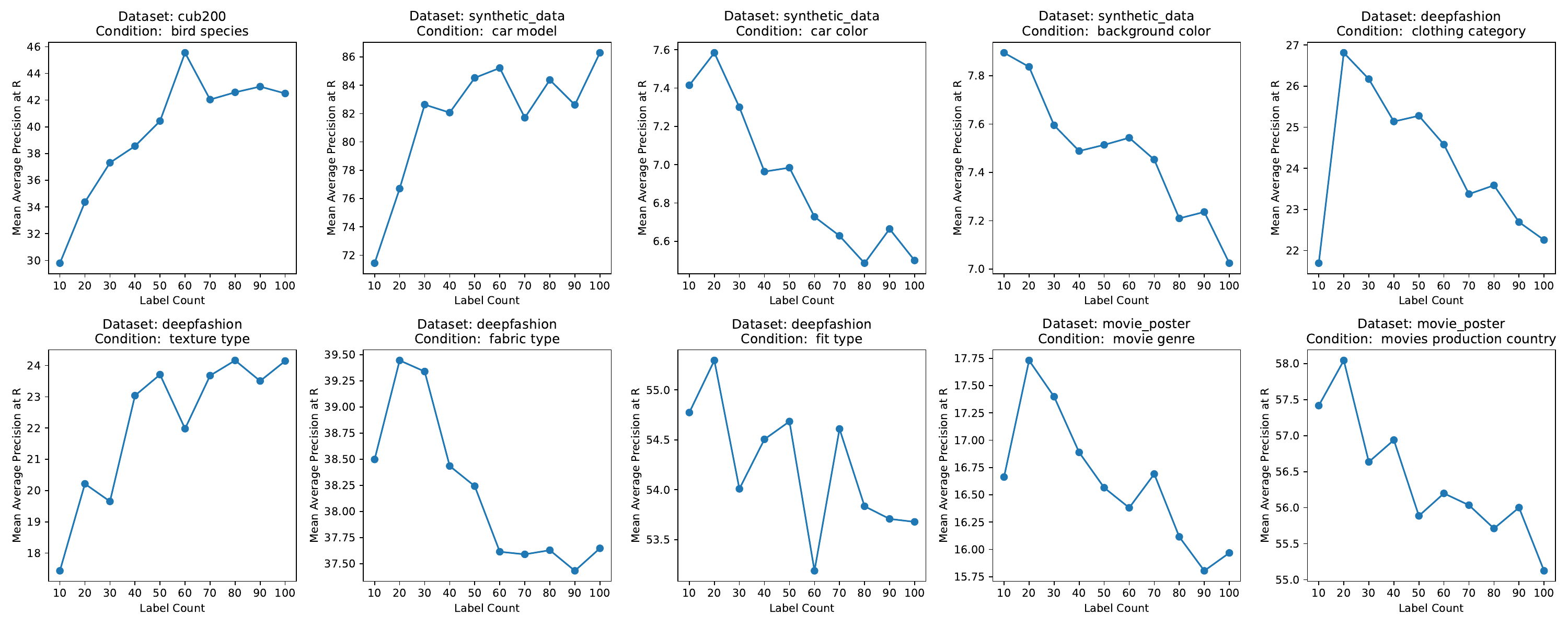}
\caption{The MAP@R score for each number of generated labels in InDiReCT.}
\label{fig:mapr_for_label_count}
\end{figure*}

\setlength{\tabcolsep}{4.5pt}
\begin{table*}[t]
\centering
\caption{Evaluation results for the LanZ-DML benchmark. Since we could not obtain the text labels from the metadata of Cars196, they are marked as N/A in InDiReCT. 
\red{* indicates that the vision encoder is not CLIP-L/14, $\dagger$ denotes methods that require training, and $\ddagger$ denotes methods that require prior knowledge of Oracle labels included in each condition.}
Avg represents the average over the respective conditions for the Synthetic Cars, Cars196~(Cars), CUB200~(CUB), DeepFashion, and Movie Posters datasets.}
\begin{tabular}{@{}lrcccccccccccc@{}}
\toprule
\multicolumn{1}{c}{} & \multicolumn{1}{c}{}
& \multicolumn{3}{c}{Synthetic Cars}
& \multicolumn{1}{c}{Cars}
& \multicolumn{1}{c}{CUB}
& \multicolumn{4}{c}{DeepFashion}
& \multicolumn{2}{c}{Movie Posters} & \\

\cmidrule(lr){3-5}\cmidrule(lr){6-6}\cmidrule(lr){7-7}\cmidrule(lr){8-11}\cmidrule(lr){12-13}
& Params
& Mod. & Col. & BG
& Mod.
& Spe.
& Clo. & Tex. & Fab. & Fit
& Gen. & Cou. & Avg\\
\midrule
CLIP  & 428M & 72.3 & 5.8 & 6.8 & 43.0 & 33.3 & 12.1 & 18.4 & 33.9 & 53.0 & 12.2 & 50.7 & 31.0 \\
EVA-CLIP* & 18B & 91.9 & 5.9 & 6.8 & 67.7 & 60.2 & 15.3 & 18.8 & 35.9 & 53.5 & 10.1 & 49.7 & 37.8 \\
\midrule
BLIP &  450M   & 22.5 & 10.2 & { 7.6} & { 9.9} & 15.5 & 11.1 & 14.5 & 33.3 & 54.3 & { 7.3} & 48.3 & 19.5 \\
BLIP2 &   4B    & 42.2 & { 5.2} & { 7.7} & 27.7 & { 6.1} & 11.7 & 16.9 & 35.3 & 54.3 & { 8.0} & 51.3 & 24.2 \\
\midrule
\textcolor{gray}{InDiReCT~(CLIP)$^\dagger$$^\ddagger$} & \textcolor{gray}{428M} &
\textcolor{gray}{94.6} &
\textcolor{gray}{7.3} &
\textcolor{gray}{7.8} &
\textcolor{gray}{N/A} &
\textcolor{gray}{44.0} &
\textcolor{gray}{24.9} &
\textcolor{gray}{24.9} &
\textcolor{gray}{40.1} &
\textcolor{gray}{54.4} &
\textcolor{gray}{17.9} &
\textcolor{gray}{55.6} & \textcolor{gray}{37.1} \\

InDiReCT (CLIP)$^\dagger$ & 428M & 86.3 & 7.6 & 7.9 & N/A & 45.5 & 26.8 & 24.2 & 39.4 & 55.3 & 17.7 & 58.0 & 36.9 \\
\textcolor{gray}{InDiReCT~(EVA-CLIP)*$^\dagger$$^\ddagger$} &
\textcolor{gray}{18B} &
\textcolor{gray}{96.7} &
\textcolor{gray}{7.7} &
\textcolor{gray}{8.0} &
\textcolor{gray}{\text{N/A}} &
\textcolor{gray}{67.1} &
\textcolor{gray}{28.9} &
\textcolor{gray}{24.9} &
\textcolor{gray}{41.5} &
\textcolor{gray}{55.0} &
\textcolor{gray}{15.2} &
\textcolor{gray}{55.2} &
\textcolor{gray}{40.0} \\

InDiReCT (EVA-CLIP)*$^\dagger$ & 18B & \textbf{94.0} & 7.5 & 7.8 & \text{N/A} & \textbf{66.9} & 32.1 & 24.0 & 40.6 & 54.9 & 14.6 & 52.3 & 39.5 \\

\midrule
CapEmb (SimCSE)  & 11B& 51.2 & 11.5 & 11.5 & 28.6 & 24.4 & 23.0 & 13.9 & 39.4 & 52.4 & 18.3 & 53.6 & 29.8 \\
CapEmb (SentenceT5)& 11B& 49.6 & 14.8 & 12.7 & 24.8 & 23.1 & 25.7 & 14.6 & 39.2 & 52.4 & 17.9 & 49.9 & 29.5 \\
\midrule
DIOR (LLaVA-1.6)& 7B & 73.7 & 13.9 & 15.6 & 40.6 & 21.7 & 26.7 & 26.6 & 42.8 & 54.5 & 20.3 & 58.1 & 35.9 \\
DIOR (LLaVA-1.6)& 13B & 78.9 & 14.9 & 15.4 & 45.8 & 23.1 & 30.8 & \textbf{34.8} & 43.4 & 57.4 & 21.9 & 60.0 & 38.8 \\
DIOR (Llama-3.2) & 11B & 88.0 & 14.7 & \textbf{17.8} & 67.3 & 51.9 & 33.7 & 26.0 & 46.4 & 57.8 & 23.4 & 62.2 & 44.5 \\
DIOR (Qwen-2.5-VL)* & 3B & 71.0 & 14.4 & 15.3 & 65.5 & 57.7 & 42.7 & 31.2 & 43.6 & 58.4 & 21.5 & 61.1 & 43.9 \\
DIOR (Qwen-2.5-VL)* & 7B & 73.2 & \textbf{15.0} & 16.8 & \textbf{68.0} & 58.1 & \textbf{45.3} & 33.9 & \textbf{47.6} & \textbf{62.1} & \textbf{28.1} & \textbf{66.0} & \textbf{46.7} \\
\bottomrule
\end{tabular}
\label{tab:result_lanz_dml_transposed_X}
\end{table*}

InDiReCT extract features from CLIP’s image embeddings using a PCA-like approach with a predefined list of labels as conditions.
However, since obtaining such a label list is not feasible in practice, we generate labels using a LLM. Specifically, we use GPT-4.1 with the following prompt to generate a list of labels: ``\texttt{Generate \{num\_generate\} labels in English to classify \{condition\} and output them in comma-separated format:"}''
Here, \texttt{\{num\_generate\}} represents the number of labels to generate.
We generate labels in increments of 10, ranging from 10 to 100.
\cref{fig:mapr_for_label_count} presents the MAR@R scores for different numbers of labels, and we report the maximum value among them. 
We denote the case where the correct labels from the dataset are used as InDiReCT~(Oracle), while the case where the generated labels are used is reported as InDiReCT~(Real).

\red{
\paragraph{BLIP/BLIP-2 Baselines}
To further contextualize the performance of our method, we included two additional vision-language models, BLIP and BLIP-2, as baselines in our LanZ-DML benchmark experiments.
For the BLIP baseline, we used the ``\texttt{blip-itm-large-coco}'' model.
The image was processed by the vision encoder, while the textual condition was fed into the text encoder. 
We then extracted the \texttt{[CLS]} token from the final multimodal output to serve as the conditional image embedding. For BLIP-2, we employed the ``\texttt{blip2-flan-t5-xl}'' model. 
In this setup, the image was processed by the vision encoder while the textual condition was passed to the text encoder, and we then used the averaged output from the Q-Former as the final embedding.

As shown in the full results in \ref{tab:result_lanz_dml_transposed_X}, both BLIP and BLIP-2 exhibit significantly lower performance on this conditional retrieval task compared to CLIP-based methods and DIOR. 
This outcome is likely due to fundamental differences in their training objectives and architecture. 
BLIP and BLIP-2 are primarily optimized for image-text matching (ITM), a task that predicts whether an image-text pair is correctly matched. This objective does not inherently cultivate a well-structured cosine similarity space for nuanced image-to-image comparisons, placing them at a disadvantage against models like CLIP. Furthermore, the Q-Former in BLIP-2 is designed to produce a compact representation to interface with a Large Language Model (LLM), which may result in a lower-resolution embedding that is less effective for standalone similarity tasks.
}

\section{GeneCIS Experiments}
\label{sec:genecis}
\paragraph{Settings}
GeneCIS~\cite{vaze2023genecis} is a benchmark that retrieves images that match the query image and condition, with objects and attribute types as important dimensions for image similarity.
Although this includes two types of tasks, including ``focus'' and ``change,'' we work on the ``focus'' task, which is consistent with our goals.
The benchmark is built on existing public datasets, such as COCO~\cite{lin2014microsoft,kirillov2019panoptic} and VAW~\cite{pham2021learning} based on Visual Genome~\cite{krishna2017visual}.
For each query, 10 to 15 images are included, one of which is the correct answer, and the evaluation metric is Recall at k (Recall@k) for k=1, 2, and 3.
For baselines, we compare against a Combiner model~\cite{baldrati2022conditioned}, which is trained with contrastive learning to combine image and text information.
We report results from two versions of the Combiner trained on different datasets: one on CIRR~\cite{liu2021image} and another on CC3M~\cite{sharma-etal-2018-conceptual}.
Furthermore, following the prior study, we include a CLIP baseline with three variants: (1) using the embedding of the text condition from CLIP text encoder (Text-only), (2) using the embedding of the query image from CLIP image encoder (Image-only), and (3) using an averaged embedding of them (Text+Image).

\input{tables/results_genecis}

\paragraph{Results}
We compare DIOR against methods that require task-specific training on the GeneCIS benchmark.
As shown in \cref{tab:result_genecis}, DIOR achieves strong performance despite not being trained specifically for this task.
This success is likely due to the effective use of the LVLM's inherent knowledge to correctly reference attributes and objects within the images.

\section{Analysis}
\label{sec:peformance_degration}
\paragraph{Performance Degradation in Highly Specific Domains}

\begin{table}[t]
\centering
\caption{Results of text generation using DIOR prompts on DomainNet and WikiArt. GT stands for Ground Truth.}
\includegraphics[width=\linewidth]{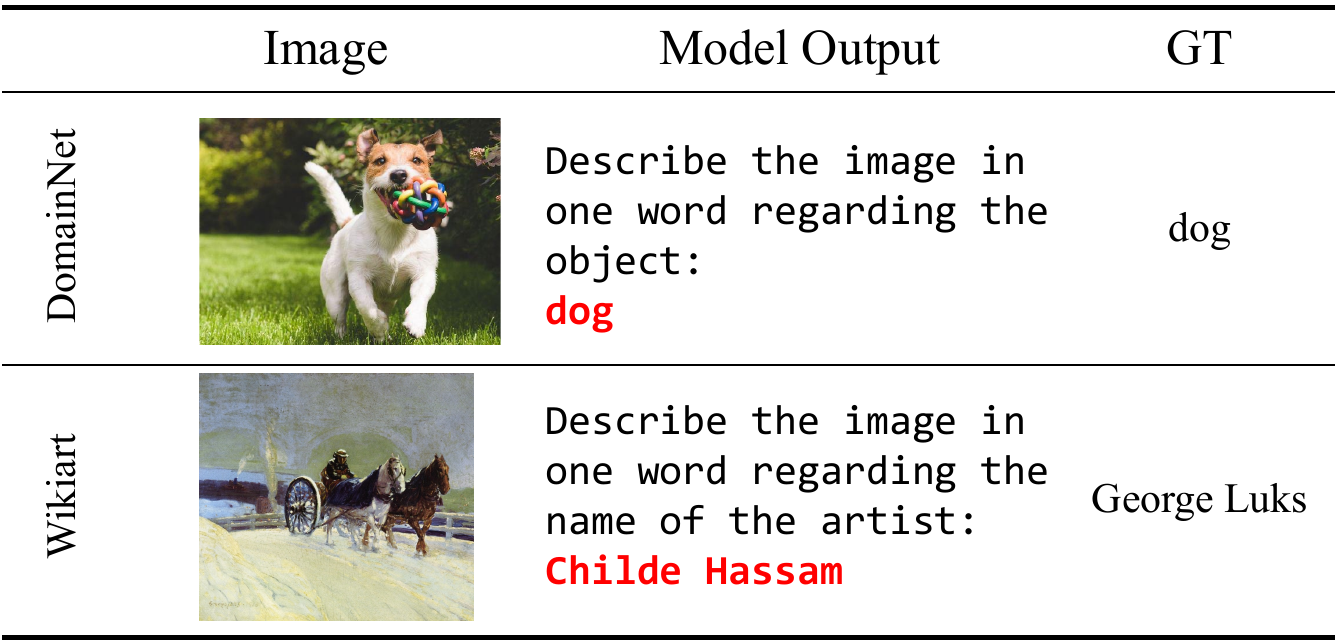}
\label{table:compare_output}
\end{table}

To investigate the performance degradation observed specifically in the Wikiart dataset, as shown in \cref{table:compare_output}.
For the DomainNet dataset, the output was ``dog,'' correctly matching the ground truth.
However, for the Wikiart dataset, ``\textit{Childe Hassam}'' is simply the wrong output. We can at least confirm that the decoded tokens are not corrupted, but the model confidently provides the wrong answer.
We attribute this failure to the LVLM's lack of profound knowledge in a specific domain.
Since LVLMs are applied across various tasks, visual instruction tuning involves training with images and responses from various domains.
DomainNet, with its diverse image types, closely aligns with the photos typically used in visual instruction tuning. 
In contrast, Wikiart represents a specialized art domain that demands deep, nuanced knowledge for accurate image interpretation. 
We conjecture that our base LVLM may lose such knowledge during construction.
Additionally, fine-tuning LVLMs for particular domains is a promising approach, which we will leave for future research.

\blue{
\paragraph{Evaluation on Image Classification Tasks}
\input{tables/results_linear_probe}

To further assess the quality of the generated, we conducted additional experiments on standard image classification tasks. 
We evaluated the performance of our embeddings on linear probing and few-shot classification using the Cars196 and CUB200 datasets, featuring a single condition (``Car Model'' and ``Bird Species,'' respectively).
For this evaluation, we compare three methods: (1) the CLIP ViT-L/14 baseline, (2) our proposed DIOR framework with the conditional prompt using Llama-3.2 (referred to as DIOR (Ours)), and (3) an ablation variant of DIOR that uses a non-conditional prompt (``Describe the image in one word:'') to investigate the impact of conditioning, referred to as DIOR (w/o condition).

The results are presented in \cref{tab:classification_results}.
Our proposed DIOR method consistently and significantly outperforms the strong CLIP baseline across all metrics in both linear probe and few-shot settings. For instance, on Cars196, DIOR achieves a linear probe accuracy of 80.02\% (k=1), a substantial improvement over CLIP's 59.82\%. 
This demonstrates that the conditional embeddings generated by DIOR are not only effective for retrieval but also produce a more linearly separable feature space, indicating higher-quality representations.

Crucially, the performance of DIOR without the conditional prompt (DIOR w/o condition) drops significantly, falling even below the CLIP baseline. This result strongly validates our core hypothesis: the performance gain is not merely due to using a larger LVLM, but is fundamentally driven by the \textit{conditional} nature of our prompting strategy. 
This highlights the effectiveness of the DIOR framework in steering the LVLM to focus on specific, relevant attributes and generate highly discriminative embeddings.
}

\section{Case Study}

\begin{figure*}[t]
\small
\centering
\includegraphics[width=0.9\linewidth]{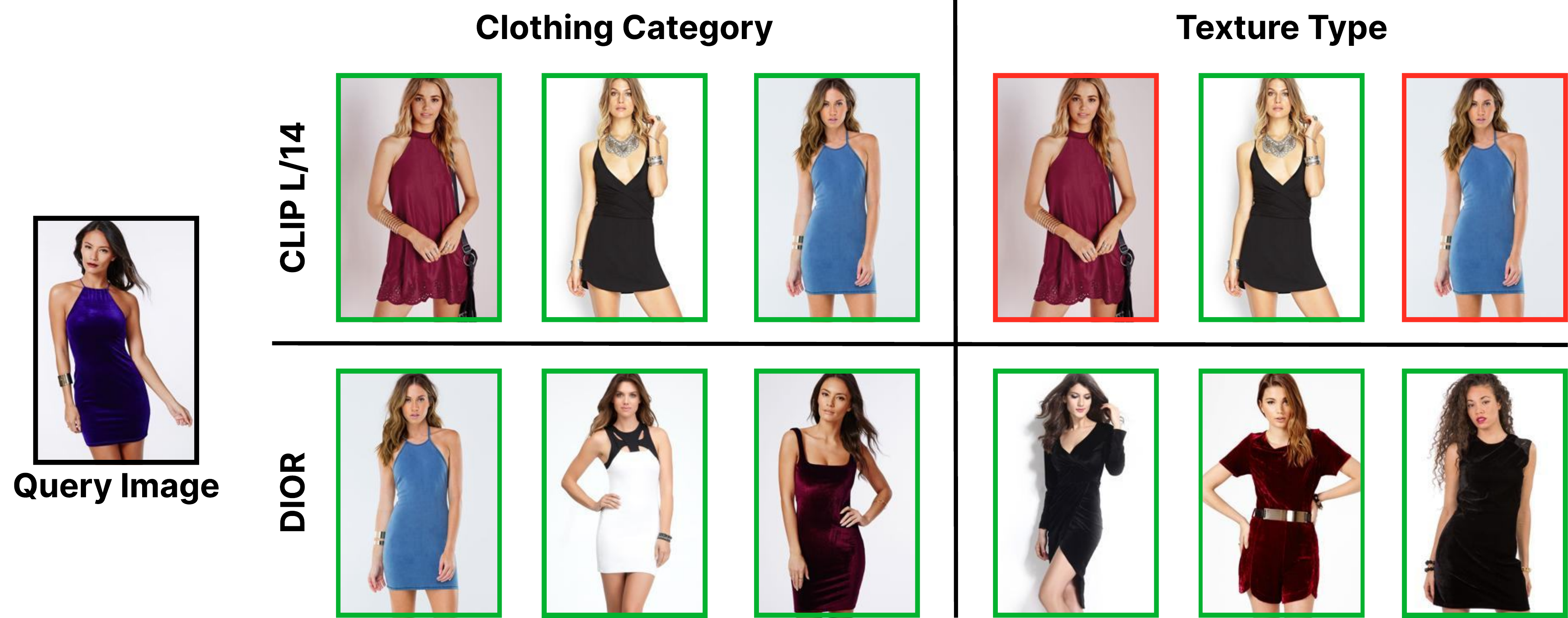}
\caption{Examples of images retrieved by CLIP L/14 and DIOR in terms of `Clothing Category' and `Texture Type' for the query image in DeeepFashion dataset.
Among the retrieved images, the ones with green boxes indicate correct cases and the ones with red boxes indicate incorrect cases.}
\label{fig:compare_retrieval}
\end{figure*}

Retrieval results on DeepFashion obtained through CLIP and DIOR are shown in \cref{fig:compare_retrieval}.
CLIP can find the correct image similar to the query image for the clothing category, but it fails for the texture type.
This suggests that image embeddings of CLIP are primarily focused on the clothing category features. 
In contrast, DIOR demonstrates appropriate retrieval in both condition.
This highlights potential advantages of DIOR in tasks that require more detailed feature distinction.

%% file: tables/results_genecis.tex
\begin{table}[t]
\centering
\caption{Evaluation Results on the GeneCIS benchmark.}
\setlength{\tabcolsep}{4pt}
\begin{tabular}{@{}lcccccc@{}}
\toprule
& \multicolumn{3}{c}{Focus Attribute} & \multicolumn{3}{c}{Focus Object} \\
& \multicolumn{3}{c}{(Recall@k)} & \multicolumn{3}{c}{(Recall@k)} \\
\cmidrule(lr){2-4} \cmidrule(lr){5-7}
Method & 1 & 2 & 3 & 1 & 2 & 3 \\
\midrule
CLIP (Text only) & 10.2 & 20.5 & 29.6 & 6.5 & 16.8 & 22.4 \\
CLIP (Image only)  & 17.7 & 30.9 & 41.9 & 9.3 & 18.2 & 26.2 \\
CLIP (Text + Image)  & 15.6 & 26.3 & 37.1 & 10.8 & 21.0 & 31.2 \\
Combiner~(CIRR) & 15.1 & 27.7 & 39.8 & 13.5 & 25.4 & 36.7 \\
Combiner~(CC3M) & 19.0 & 31.0 & 41.5 & 14.7 & 25.9 & 36.1 \\
DIOR            & \textbf{24.0} & \textbf{36.8}  & \textbf{47.0}  & \textbf{21.1} & \textbf{34.3} & \textbf{42.9} \\
\bottomrule
\end{tabular}
\label{tab:result_genecis}
\end{table}

%% file: tables/results_linear_probe.tex
\begin{table}[t]
\centering
\caption{Linear probe and few-shot classification accuracy (\%) on the Cars196 and CUB200 datasets. Our proposed method, DIOR (Ours), significantly outperforms both the CLIP baseline and its non-conditional variant, highlighting the effectiveness of our conditional prompting strategy. Best results are in \textbf{bold}.}
\label{tab:classification_results}
\resizebox{\columnwidth}{!}{%
\begin{tabular}{@{}lcccccc@{}}
\toprule
\textbf{Cars196} & \multicolumn{3}{c}{\textbf{Linear Probe}} & \multicolumn{3}{c}{\textbf{Few-shot}} \\
\cmidrule(lr){2-4} \cmidrule(lr){5-7}
Method & k=1 & k=5 & k=10 & k=1 & k=5 & k=10 \\
\midrule
CLIP ViT-L/14 & 59.82 & 79.41 & 85.38 & 55.41 & 81.18 & 86.04 \\
DIOR~(w/o condition) & 40.06 & 58.83 & 69.53 & 39.44 & 64.14 & 72.43 \\
DIOR~(Ours) & \textbf{80.02} & \textbf{89.67} & \textbf{93.15} & \textbf{77.47} & \textbf{91.33} & \textbf{93.38} \\
\bottomrule
\\
\toprule
\textbf{CUB200} & \multicolumn{3}{c}{\textbf{Linear Probe}} & \multicolumn{3}{c}{\textbf{Few-shot}} \\
\cmidrule(lr){2-4} \cmidrule(lr){5-7}
Method & k=1 & k=5 & k=10 & k=1 & k=5 & k=10 \\
\midrule
CLIP ViT-L/14 & 49.11 & 68.80 & 75.59 & 46.15 & 72.00 & 78.42 \\
DIOR~(w/o condition) & 31.28 & 46.41 & 52.78 & 33.22 & 53.77 & 60.66 \\
DIOR~(Ours) & \textbf{62.21} & \textbf{75.57} & \textbf{78.94} & \textbf{60.89} & \textbf{79.54} & \textbf{81.81} \\
\bottomrule
\end{tabular}%
}
\end{table}